\documentclass[11pt]{article}

\usepackage[preprint]{acl}

\usepackage{times}
\usepackage{latexsym}

\usepackage[T1]{fontenc}

\usepackage[utf8]{inputenc}

\usepackage{microtype}

\usepackage{inconsolata}

\usepackage{graphicx}
\usepackage{amsmath}
\usepackage{xcolor}        
\usepackage{colortbl}      
\usepackage{amssymb}       
\usepackage{array}         
\usepackage{url}           
\usepackage{makecell}
\usepackage{multirow} 

\usepackage[most]{tcolorbox}
\usepackage{amssymb}
\usepackage{tikz}
\usepackage{standalone}
\usetikzlibrary{calc, positioning, arrows.meta, bending, decorations.markings, shapes.geometric, backgrounds, shadows.blur, patterns.meta}
\usetikzlibrary{arrows.meta, positioning, shapes.geometric, shapes.misc, calc, backgrounds, shadows, fit, 3d}
\usepackage{pgfplots}
\usepackage{xcolor}
\usepackage{booktabs}

\title{Truth as a Trajectory: What Internal Representations Reveal About Large Language Model Reasoning}

\author{%
  \begin{tabular}{c}
    \textbf{Hamed Damirchi$^{1}$ \qquad Ignacio Meza De la Jara$^{1}$ \qquad Ehsan Abbasnejad$^{2}$} \\
    \textbf{Afshar Shamsi$^{3}$ \qquad Zhen Zhang$^{1}$ \qquad Javen Shi$^{1}$} \\[1ex]
    \textnormal{$^{1}$Australian Institute for Machine Learning, Adelaide University \qquad
               $^{2}$Monash University} \\ \textnormal{$^{3}$Concordia University} \\[0.5ex]
    \small{\texttt{$^{1}$\{firstname.lastname\}@adelaide.edu.au \quad $^{2}$\{firstname.lastname\}@monash.edu}} \\ 
    \small{\texttt{$^{3}$afshar.shamsi@concordia.ca}}
  \end{tabular}
}

\newtcolorbox{takeawaybox}{
  colback=blue!6,
  colframe=blue!55,
  boxrule=0.6pt,
  arc=3pt,
  left=6pt,
  right=6pt,
  top=5pt,
  bottom=5pt
}

\begin{document}
\maketitle

\begin{abstract}
Existing explainability methods for Large Language Models (LLMs) typically treat hidden states as static points in activation space, assuming that correct and incorrect inferences can be separated using representations from an individual layer. However, these activations are saturated with polysemantic features, leading to linear probes learning surface-level lexical patterns rather than underlying reasoning structures. We introduce Truth as a Trajectory (TaT), which models the transformer inference as an unfolded trajectory of iterative refinements, shifting analysis from static activations to layer-wise geometric displacement. By analyzing displacement of representations across layers, TaT uncovers geometric invariants that distinguish valid reasoning from spurious behavior. We evaluate TaT across dense and Mixture-of-Experts (MoE) architectures on benchmarks spanning commonsense reasoning, question answering, and toxicity detection. Without access to the activations themselves and using only changes in activations across layers, we show that TaT effectively mitigates reliance on static lexical confounds, outperforming conventional probing, and establishes trajectory analysis as a complementary perspective on LLM explainability.
\end{abstract}
\section{Introduction}
\begin{figure}[h]
    \centering
    \includegraphics[width=0.9\linewidth]{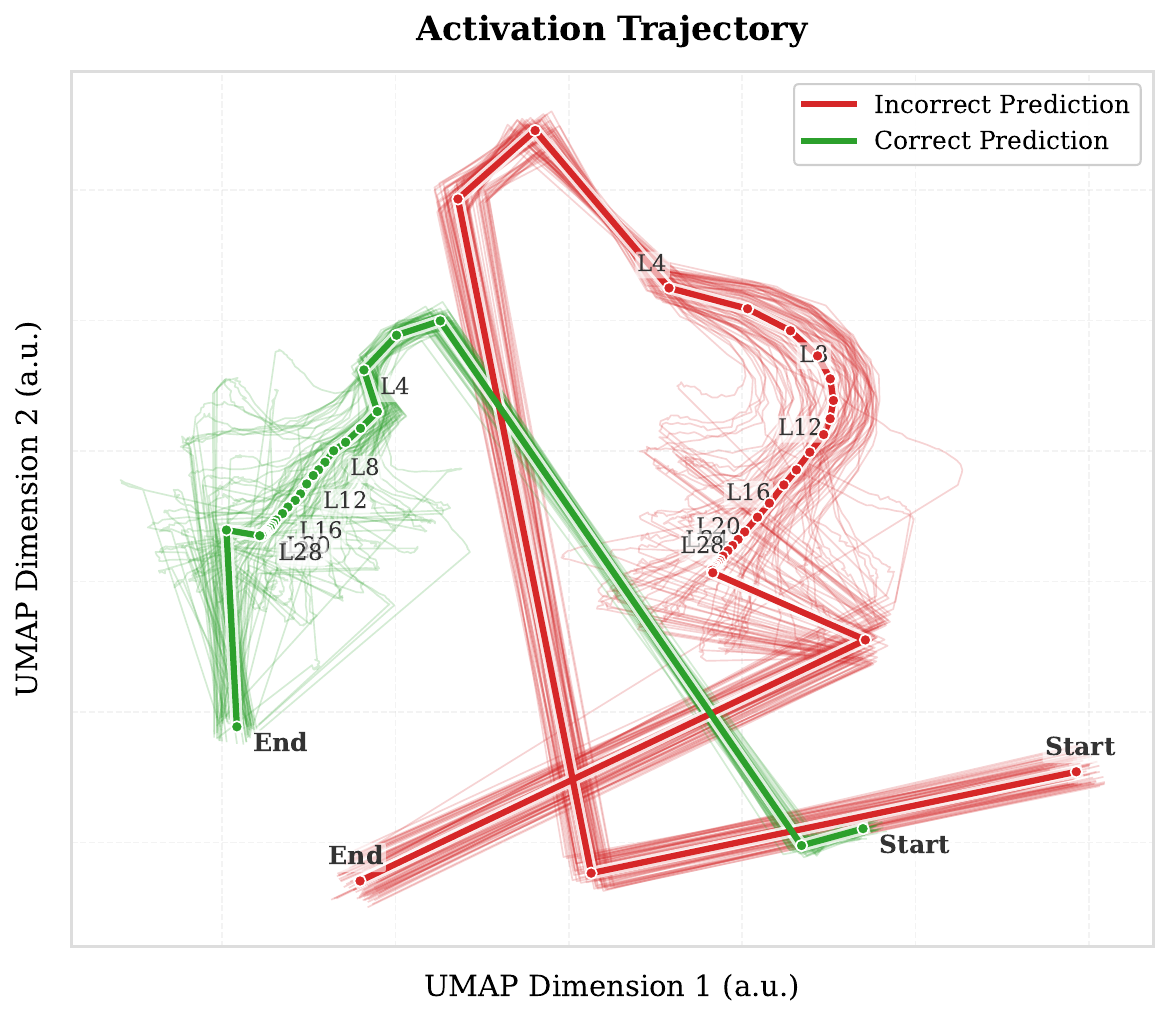}
    \caption{
    \textbf{Trajectories reveal structure beyond static embeddings.} We plot layerwise hidden states as trajectories in activation space. Correct generations (green) follow smoother paths, while incorrect ones (red) exhibit sharp deviations. Although the supervised projection amplifies separation, the geometry suggests that modeling entire trajectories, rather than isolated states, can help distinguish valid from spurious reasoning.
    }
    \label{fig:umap-hellaswag}
    \vspace{-0.6em}
\end{figure}

The deployment of Large Language Models (LLMs) in safety-critical domains, from legal reasoning to content moderation, has rendered the black-box evaluation of final outputs insufficient ~\cite{orgad2025llms,Shailya2025LExTTE,aljohani2025comprehensive}. While behavioral benchmarks measure what a model generates, they provide limited insight into how it arrives at its conclusions. Without visibility into the internal process, practitioners cannot reliably distinguish between a model that follows reasoning patterns correctly and one that merely relies on surface-level heuristics. This gap creates a fundamental bottleneck for safety. To trust these systems, we must be able to verify the validity of their internal thought processes, not just the probability of their tokens.

Recent mechanistic interpretability literature has been consistent with the Linear Representation Hypothesis \cite{park2023linear}, which suggests that high-level properties such as inference validity or toxicity are encoded as distinct linear directions within the model's activation space. Thus, interpretability and behavior editing approaches assume certain properties can be classified via linear probes \cite{bao-etal-2025-probing} or manipulated via activation steering \cite{aljohani2025comprehensive,rimsky-etal-2024-steering,anonymous2025enhancing,obrien2025steeringlanguagemodelrefusal} at carefully chosen layers. Under this view, explainability reduces to identifying the "right" static layer and direction that separates valid from invalid behavior. A limitation of these approaches is their reliance on contrastive samples for the targeted property. Consequently, the efficacy of probing and steering results becomes subject to the specific datasets employed because they operate on static activations.

Thereby, despite early optimism, recent evidence suggests that the "geometries of truth" are often task-specific and orthogonal across domains \cite{azizian2025the}, i.e., a probe trained to detect inference correctness in one context fails to generalize beyond the distribution it was trained on. This failure may be driven by the polysemantic nature of transformer activations \cite{lindsey2025biology}, which simultaneously encode lexical content, syntactic structure, and task-specific artifacts. This way, linear probes may latch onto surface-level correlations, such as the presence of specific tokens, rather than the underlying validity of the reasoning patterns. Furthermore, the selection of which layer to probe remains unprincipled. Findings from the activation steering literature indicate that effective intervention is only possible in a narrow mid-layer section of LLMs with inconsistent results across datasets \cite{rimsky-etal-2024-steering}, while other works suggest that the reasoning manifold is fundamentally non-linear \cite{manson2025curved}, resisting simple linear separation. These inconsistencies are often a byproduct of evaluations on synthetic or tightly controlled datasets, leaving open the question of whether a generalizable, task-agnostic signature of reasoning validity exists in real-world benchmarks.

We propose \textbf{Truth as a Trajectory (TaT)}, which reframes LLM inference as a dynamical process rather than a collection of static layer snapshots. TaT unfolds the inference pass across \emph{layers and tokens} into a trajectory through representation space (see Figure \ref{fig:umap-hellaswag}). We shift the analysis from raw activations to layer-wise displacements, i.e., the difference between successive residual stream activations across layers. This transformation mitigates reliance on static token-identity and lexical features, isolating how representations are updated across depth rather than what is encoded at a single layer. To capture the dynamics of this evolution, we use a lightweight LSTM classifier. While we explored simpler kinematic measures such as velocity, acceleration, and curvature inspired by recent work on transformer dynamics \cite{zhou2025geometry,fernando2025transformerdynamics}, we found them to be inconsistent predictors of reasoning validity across diverse tasks. Our learned approach, on the other hand,  captures the non-linear structural invariants associated with valid reasoning. To our knowledge, this is the first approach to model the "internal thought process" of an LLM on the reasoning validity of the input statement by unfolding the activations of every token across all layers into a continuous trajectory. 

We evaluate TaT on widely used benchmarks spanning commonsense reasoning, question answering, factuality, and toxicity detection, across both dense and Mixture-of-Experts (MoE) architectures. Our main finding is that TaT achieves strong cross-dataset generalization. A trajectory classifier trained on a single source dataset generalizes across tasks with varying task-prompt structure, outperforming both linear probing baselines and the base model's own zero-shot and In-Context Learning (few-shot) performance on different domains. This suggests that the trajectory of valid reasoning encodes structural invariancies that transcend task-specific lexical patterns. On toxicity detection, TaT is more robust to lexical confounds, e.g., it can better distinguish quoted or contextualized toxic vocabulary from toxic intent and consistently outperforms probes trained on mid-layer or final-layer activations. Our contributions are summarized below:

\begin{itemize}
    \item \textbf{Trajectory-based explainability}: We introduce Truth as a Trajectory (TaT), which models LLM inference as a dynamical process that unfolds across layers and tokens, capturing the continuous geometric evolution of reasoning rather than focusing on individual layers.
    \item \textbf{Cross-task Geometric Invariants}: By analyzing layer-wise displacement vectors rather than activations themselves, we mitigate reliance on static lexical features and emphasize the model's internal geometric refinement process, exposing trajectory-level structure that is unobservable to linear probes.
    \item \textbf{Trajectory-based behavior detection}: We demonstrate that trajectory analysis can be extended to complex behavioral properties, such as toxicity. TaT significantly outperforms linear probes in distinguishing meaningful context from toxic intent, validating its utility for reliable model monitoring.
\end{itemize}
\section{Related Work}
\label{sec:related_work}

Mechanistic interpretability has traditionally focused on static linear representations of concepts within isolated layers. However, the residual stream structure of Transformers suggests a dynamical systems perspective, where the evolution of representations across layers may provide insights into the thought process of the model. We situate our work at the intersection of these paradigms, focusing on the geometry of the inference trajectory.

\subsection{Static Linear Representations}
The current predominant paradigm in mechanistic interpretability is the \textit{Linear Representation Hypothesis} (LRH), which suggests that neural networks represent high-level concepts as linear directions in activation space \cite{park2023linear, elhage2022toy, liu2025ipredictiam}. This view has motivated works on classifying behaviors via linear probes \cite{belinkov2022probing} or interpreting them via Sparse Autoencoders (SAEs) \cite{cunningham2023sparse}. However, these methods face significant limitations. Both require exhaustive layer-wise searches, as the specific depth of concepts is unknown \textit{a priori}, and SAEs notably lack consistent structural mapping across models, complicating cross-model interpretation. Furthermore, linear probing relies on subjective, curated datasets to define target behaviors, making the discovered directions dependent on specific semantic content rather than intrinsic model geometry. While extensions like Contrast-Consistent Search (CCS) \cite{burns2022discovering} attempt to find latent knowledge without supervision they, along with standard probing, typically analyze representations as static vectors within isolated layers (e.g., "layer $L$"). This ignores the \textit{temporal evolution} of the inference process. Our work challenges this static, content-dependent view, arguing that reasoning validity is a dynamic property best captured by the geometric displacement of activations across the computational trajectory, rather than their static positioning in fixed coordinates.

\subsection{Representation Engineering and Steering}
Representation Engineering (RepE) shifts focus from analysis to control. \citet{zou2023representation} demonstrated that extracting "concept vectors" allows for top-down steering of model behavior, inhibiting hallucinations or toxic outputs by injecting these vectors into the residual stream. However, static steering is not guaranteed to predictably steer towards desired behavior for every model and task \cite{rimsky-etal-2024-steering}. This is potentially due to some datasets requiring the model's internal state to navigate a non-linear path. Thus, recent adjacent literature advocates in favor of alternative approaches. \citet{zhang2025empirical,manson2025curved} argue that simple linear interventions cannot account for the "manifold evolution" during multi-step inference. Similarly, \citet{postmus2024steering} suggests that steering should target activation regions via transformations rather than static directions. Meanwhile, like linear probing, these methods remain dependent on subjective, curated datasets to define target behaviors. Our approach mitigates these issues by analyzing the displacement of activations across the entire trajectory, capturing how activation trajectories behave in task-specific regions without relying on layer-wise activations.

\subsection{Transformers as Dynamical Systems}
The theoretical interpretation of Transformers as discretized dynamical systems provides the basis for our kinematic framework. The residual update $\mathbf{h}_{\ell+1} = \mathbf{h}_\ell + f(\mathbf{h}_\ell)$ is mathematically equivalent to a step of the Euler method for solving an Ordinary Differential Equation (ODE) \cite{chen2018neural, lu2019understanding}. This equivalence implies that the layer-wise evolution of activations can be analyzed as a continuous trajectory in a high-dimensional state space, rather than as discrete, independent states. \citet{geshkovski2023transformers} extended this view to self-attention, modeling tokens as interacting particles that converge toward semantic clusters over "time" (depth). We conduct an extensive study on kinematic measures of activation trajectories across various models and datasets. We demonstrate that while pre-determined measures lack universal applicability, the intrinsic motion of internal representations may contain useful signals. Consequently, we model this motion through the activation space, showing that this trajectory can accurately classify the validity of the input statement's reasoning process or behavior characteristics.

\subsection{The Geometry of Inference}
Most relevant to our proposal is the emerging body of work on the "geometry of reasoning". \citet{zhou2025geometry} recently proposed that logical validity governs the "velocity field" of the representation flow, while semantic content determines position. They demonstrate that logical deductions trace specific flow patterns distinct from semantic association. However, their analysis remains largely theoretical, focusing on idealized logical structures rather than the noisy, unstructured data typical of real-world LLM usage. Concurrently, \citet{manson2025curved} introduced the concept of "Curved Inference," showing that "semantic concern" (e.g., urgency or moral framing) induces measurable curvature in the residual stream. This finding emphasizes the inadequacy of static linear probes, which often conflate these geometric nuances with semantic content. We demonstrate that these kinematic signatures are not only present but \textit{learnable} in general settings. Unlike prior works restricted to curated datasets, we leverage these geometric trajectories to distinguish between correct and incorrect behavior in complex, real-world benchmarks.

\section{Problem Setup}
\label{sec:setup}
We consider a standard evaluation setting where a model is presented with a context or prompt $\mathbf{x}$ and a set of candidate continuations $\mathcal{C} = \{c_1, c_2, \dots, c_k\}$. These continuations may range from single tokens (e.g., "True"/"False") to complete sentences or reasoning chains. For each candidate $c_i$, we construct a complete input sequence by concatenating the prompt with the continuation. We then perform a forward pass through the Transformer model to extract the internal representations.

Let $L$ denote the number of layers in the model and $N_i$ be the number of tokens in the concatenated sequence for candidate $c_i$. As the input is processed, each transformer block updates the residual stream, producing a sequence of activation vectors. We collect the activations from the output of every transformer block for all tokens to form a trajectory matrix $\mathcal{T}_i \in \mathbb{R}^{M_i \times d}$, where $d$ is the hidden dimension size and $M_i = N_i \times L$ represents the total vectors in the unrolled computation graph. 

Our primary objective is to analyze the geometric properties of these trajectories. Specifically, we aim to determine whether the trajectory $\mathcal{T}_i$ corresponding to the correct continuation exhibits distinct kinematic signatures compared to the trajectories of incorrect candidates.

\section{Geometry of Inference}
\label{sec:theory}

Our goal in this section is to kinematically analyze the activation trajectory extracted using the process described in section \ref{sec:setup}. This investigation is motivated by a notable gap between theoretical work on transformer dynamics and practical activation probing. While recent studies on the dynamics of activation spaces have uncovered intriguing properties, e.g., characteristic velocity profiles in idealized settings, these insights are rarely tested against standard, diverse benchmarks. Conversely, probing methods often ignore the temporal evolution of inference. Here, we determine whether simple, interpretable kinematic measures can distinguish between correct and incorrect reasoning patterns in complex scenarios. We employ an oracle-guided analysis where the ground truth is known a priori. By distinguishing between correct and incorrect outputs, we aim to identify \emph{consistent} kinematic rules, such as differences in statistics of velocity, curvature, acceleration, etc., that consistently correlate with correct reasoning patterns.

\subsection{Kinematic Descriptors}

For a fixed input, a Transformer with $L$ layers produces a sequence of hidden states $\mathcal{T} = (h_0, h_1, \dots, h_L)$, where each layer applies a residual update as follows:
\begin{equation}
    h_{\ell+1} = h_\ell + f_\ell(h_\ell).
\end{equation}
This update can be interpreted as a discrete-time evolution of the hidden state across depth. This way, layerwise kinematic descriptors can be derived from $\mathcal{T}$ that capture magnitude, effort, and directional consistency.
We define the displacement vector at layer $\ell$ as $\Delta h_\ell := h_{\ell+1} - h_\ell$. This vector represents the update applied by the $\ell$-th transformer block to the residual stream.

\paragraph{Velocity}
Velocity is defined as $v_\ell := \|\Delta h_\ell\|_2$ and measures the magnitude of changes in activation vectors between consecutive layers. Note that our goal is not to interpret what a large or small velocity is, but rather to compare velocity profiles between correct and incorrect generations and identify consistent patterns across models and datasets in an unbiased manner.

\paragraph{Acceleration}
Acceleration is defined as $a_\ell := v_\ell - v_{\ell-1}$, i.e. the gradient of velocity. 

\paragraph{Jerk}
Jerk is defined as the rate of change of acceleration, $j_\ell := a_\ell - a_{\ell-1}$ and captures the smoothness of the trajectory's evolution. 

\paragraph{Directional Curvature}
Curvature measures directional consistency between successive updates:
\begin{equation}
    \kappa_\ell :=
    \frac{\langle \Delta h_\ell, \Delta h_{\ell-1} \rangle}
    {\|\Delta h_\ell\|_2 \, \|\Delta h_{\ell-1}\|_2}.
\end{equation}
This formulation captures the angular deviation between consecutive displacements, with higher curvature indicating more abrupt directional changes. 

\paragraph{Kinematic Curvature}
While directional curvature captures angular deviation, we also define a geometric curvature metric based on the instantaneous kinematics of the trajectory. Let $\mathbf{v}_\ell$ and $\mathbf{a}_\ell$ denote the vector velocity and acceleration, respectively. The kinematic curvature is defined as:
\begin{equation}
    \kappa^{\text{kin}}_\ell := \frac{\|\mathbf{a}_\ell\|_2}{\|\mathbf{v}_\ell\|_2^2}.
\end{equation}
This quantity becomes large when the trajectory exhibits large acceleration relative to its step size (i.e., when $\|\mathbf{v}_\ell\|$ is small), capturing abrupt changes in the update direction even when the overall movement through activation space is modest.

\paragraph{Arc Length}
Arc length quantifies the total distance traversed by the activation vector as it evolves through the network depth:
\begin{equation}
    S := \sum_{\ell=0}^{L-1} \|h_{\ell+1} - h_\ell\|_2.
\end{equation}
This metric serves as a proxy for the total geometric effort exerted by the model. Note that arc length is a global descriptor and provides a scalar value for the entire trajectory.

\paragraph{Are Kinematic Descriptors Useful?}
We evaluate these descriptors on LLAMA~3.1--8B (Figure~\ref{fig:llama_kinematics}) and Qwen~2.5-14b (Figure~\ref{fig:qwen_kinematic} in Appendix \ref{app:qwen_kinematic}) using reasoning and factuality benchmarks. Kinematic signals (Velocity, Acceleration, Jerk) classify generations as \emph{correct} or \emph{incorrect}. Velocity outperforms other descriptors across datasets, in line with the suggested hypothesis in \citet{zhou2025geometry} that validity is reflected in characteristic velocity profiles distinct from semantic position. However, no single descriptor consistently matches the model's own accuracy when evaluated using normalized log-likelihood in a zero-shot setting. Extending the analysis to Qwen2.5-14B reveals even less consistency. While descriptors generally outperform a random classifier, except ARC-Easy, none approach the accuracy of the base model. To control for potential issues from our trajectory formulation, we also experimented with forming trajectories using only continuation tokens, isolating the last token's activations, and averaging activations to yield length-$L$ trajectories. In all variations, performance only degraded. We note that kinematic descriptors output scalar magnitudes. Therefore, the directional information of the activation space remains unobservable to these descriptors.
\begin{figure}[t]
    \centering
    \includegraphics[width=0.499\textwidth]{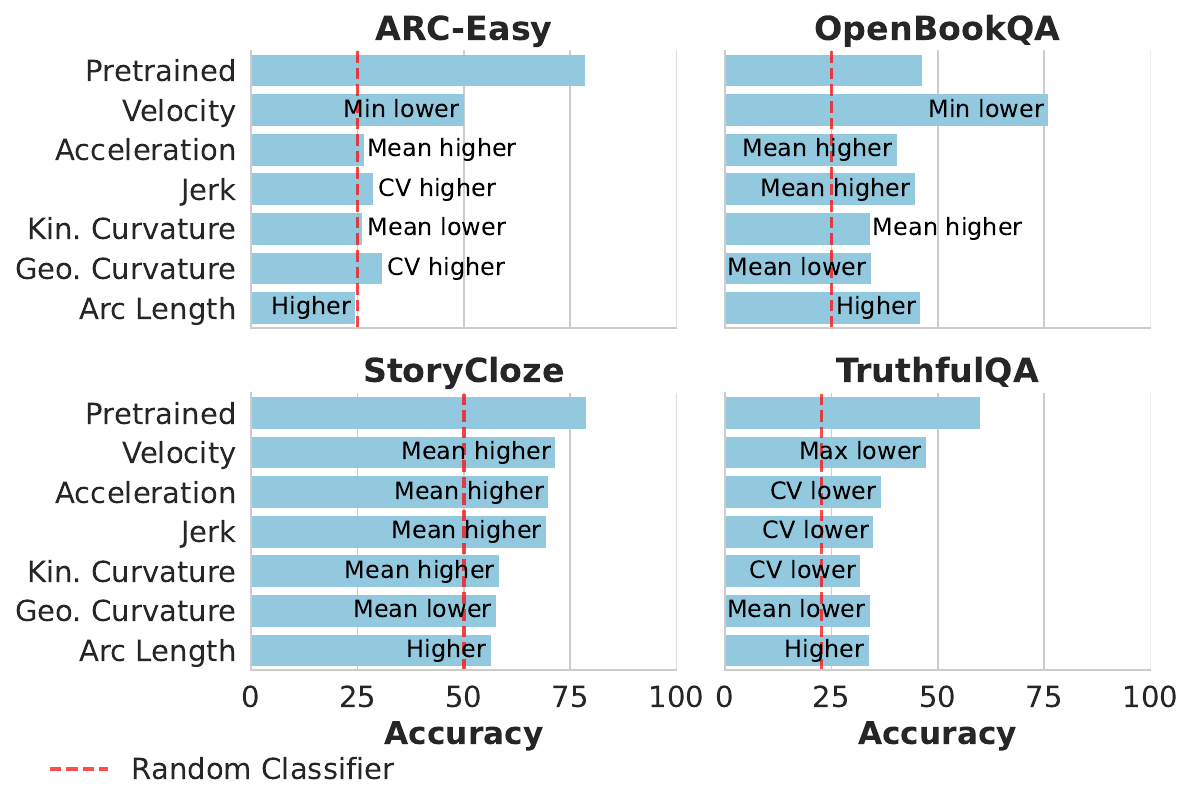}
    \caption{Performance on 4 reasoning benchmarks using kinematic descriptors. The red dashed line is the random classifier. While activation velocity obtains better results than the base model, there is no consistency in this performance improvement across datasets.}
    \label{fig:llama_kinematics}
\end{figure}
\begin{takeawaybox}
\textbf{Takeaway.}
The predictive signal in activation velocity implies that the rate and direction of the trajectory activations encode information about reasoning validity. However, static rules fail to generalize consistently. Thus, while kinematic descriptors may capture aspects of reasoning dynamics, they are insufficient. This motivates the need for learned models to better interpret these geometric signals.
\end{takeawaybox}
\subsection{Truth as a Trajectory (TaT)}
\label{sec:method}

Building on the kinematic analysis in Section~\ref{sec:theory}, we propose \textbf{Truth as a Trajectory (TaT)}, a learnable framework for detecting reasoning validity. While scalar kinematic descriptors (e.g., velocity, curvature) capture some geometric intuition, they discard the rich directional information embedded in the high-dimensional activation space. TaT addresses this by learning decision boundaries directly on the manifold of trajectory dynamics.

\subsubsection{Trajectory Construction}
Recall from Section~\ref{sec:setup} that for a candidate continuation $c_i$, we extract a trajectory of activations. To isolate the geometric evolution of reasoning from static lexical cues and other persistent content in raw activations, we transform these vectors into a sequence of layer-wise displacements.
We define the displacement vector as:
\begin{equation}
    \mathbf{d}_{t, \ell} = h_{t, \ell+1} - h_{t, \ell}
\end{equation}
We motivate this transformation through the lens of the \textit{Privileged Basis Hypothesis} \cite{elhage2023privileged, bricken2023monosemantic}. Raw activations $h_{t, \ell}$ are often dominated by high-magnitude, relatively persistent components (including token and prompt-specific content), making them susceptible to interference from polysemanticity and superposition. By taking the difference between layers, we attenuate such compounds and effectively subtract this static background, isolating the active residual update $f_\theta(h_{t, \ell})$. Due to the element-wise non-linearities in the Transformer, these updates are constrained to a consistent reference frame (Residual Alignment). Consequently, distinct from raw states which ambiguously signal if a feature is \textit{present}, the displacement $\mathbf{d}_{t, \ell}$ precisely signals if the model is \textit{actively writing} to that feature. This isolates the mechanics of the reasoning process from the semantics of the model's memory state.

We stack layer-wise updates across all tokens and layers to form a continuous trajectory sequence $\mathbf{S}_i$:
\begin{equation}
    \mathbf{S}_i = [\mathbf{d}_{1, 0}, \dots, \mathbf{d}_{1, L-1}, \mathbf{d}_{2, 0}, \dots, \mathbf{d}_{N_i, L-1}]
\end{equation}
where $\mathbf{S}_i \in \mathbb{R}^{M_i \times d}$ and $M_i = N_i \times L$ is the total trajectory length. This formulation unfolds the inference process into a single temporal sequence, treating the progression through layers and tokens as a unified path.

\subsubsection{Modeling Dynamics with LSTM}
To capture the non-linear structural invariants of valid reasoning within $\mathbf{S}_i$, we use a Long Short-Term Memory (LSTM) network. We choose an LSTM over Transformer-based probes to explicitly model the sequential dependency of the geometric updates and to maintain a lightweight computational footprint. The LSTM processes the trajectory sequence $\mathbf{S}_i$ step-by-step. After processing the entire sequence of length $M_i$, the final hidden state $\mathbf{z}_{M_i}$ encodes the geometric trajectory. This vector is passed through a linear classification head $\mathbf{W} \in \mathbb{R}^{d_{lstm} \times 1}$ to predict the probability of validity:
\begin{equation}
    \hat{y}_i = \sigma(\mathbf{W}^T \mathbf{z}_{M_i} + b)
\end{equation}
By training on the displacement trajectory $\mathbf{S}_i$, the model learns geometric signatures associated with inference correctness, generalizing beyond the specific lexical and prompt content.
\section{Experiments}
\label{sec:experiments}

We present a comprehensive evaluation of the Truth as a Trajectory (TaT) framework to assess its ability to distinguish valid reasoning from spurious correlations. Our experiments span a diverse set of domains, including commonsense reasoning, reading comprehension, factuality, and toxicity detection, across both dense (Llama-3.1-8B, Qwen2.5-14B/32B) and Mixture-of-Experts (Qwen2.5-30B MoE) architectures. We compare the performance of our trajectory-based classifier against standard static linear probing baselines and the underlying frozen language model's intrinsic zero-shot and few-shot capabilities (which we refer to as the \emph{base model}). For implementation details we refer to Appendix \ref{app:details}.

\subsection{Are Reasoning Trajectories Generalizable?}
\label{sec:universal_reasoning}

We investigate whether the geometric signature of valid reasoning is consistent across different tasks. If TaT captures a fundamental structural invariant of "truth" (or validity) rather than task-specific confounds, a trajectory classifier trained on one dataset should generalize to unseen datasets without fine-tuning on the unseen task.

\paragraph{Setup} We evaluate on a suite of reasoning benchmarks: ARC-Easy (ARC-E), ARC-Challenge (ARC-C) \cite{clark2018think}, BoolQ \cite{clark2019boolq}, Hellaswag \cite{zellers2019hellaswag}, OpenBookQA (OpenQA) \cite{mihaylov2018suit}, StoryCloze \cite{mostafazadeh2016corpus}, CommonsenseQA (ComQA) \cite{talmor2019commonsenseqa}, CosmosQA (CosQA) \cite{huang2019cosmos}, and SocialIQA (SiQA) \cite{sap2019social}. For each dataset, we train a TaT classifier (using layer-wise displacement) and a linear probe (mid layer probe) on the training split and evaluate them on all other datasets' evaluation splits. Because probe performance can be highly layer-dependent, we use the middle layer as a standard choice, and report a sweep over mid-to-late layers in Appendix~\ref{app:probe_layer_sweep}.

\begin{table*}[htbp]
    \centering
    \caption{Benchmark Accuracies across Evaluation Datasets with ID and OOD metrics. Values inside parentheses for base model accuracy indicate datasets where evaluation through normalized log-likelihood results in a degraded performance. While unfair to our approach (since this base model sees all possible answers, whereas our method does not), our approach still outperforms the base model itself in most cases.}
    \label{tab:combined_benchmarks_extended}
    \resizebox{\textwidth}{!}{%
        \begin{tabular}{llcccccccccccc}
            \toprule
            \multirow{2}{*}{\textbf{Train Dataset}} & \multirow{2}{*}{\textbf{Method}} & \multicolumn{9}{c}{\textbf{Evaluation Dataset}} & \multirow{2}{*}{\textbf{Avg}} & \multirow{2}{*}{\textbf{ID Acc.}} & \multirow{2}{*}{\textbf{OOD Avg.}} \\
            \cmidrule(lr){3-11} 
            
             & & \textbf{ARC-C} & \textbf{ARC-E} & \textbf{OpenQA} & \textbf{BoolQ} & \textbf{Hellaswag} & \textbf{CosQA} & \textbf{SiQA} & \textbf{ComQA} & \textbf{StoryCloze} & & & \\
            \midrule
            
            \multicolumn{2}{c}{Zero-shot Accuracy} & 50.1 & 78.5 & 62.4 & 74.7 & 57.4 & 81.7 (26.1) & 65.2 (48.0) & 62.4 & 78.8 & 67.9 (59.8) & - & - \\
            \multicolumn{2}{c}{Few-shot Accuracy}  & 65.3 & 84.3 & 67.0 & 83.8 & 76.8 & 82.0 (44.5) & 67.7 (60.8) & 71.1 & 83.1 & 75.7 (70.7) & - & - \\
            \midrule
            \midrule
            \multirow{2}{*}{ARC-C} 
                & Linear Probe & 75.32 & 80.09 & \textbf{78.60} & 55.48 & 70.35 & 73.34 & 65.47 & \textbf{74.59} & 66.01 & 71.03 & 75.32 & 70.49 \\
                & \textbf{TaT (Ours)}   & \textbf{82.17} & \textbf{85.31} & 73.60 & \textbf{96.91} & \textbf{73.89} & \textbf{74.24} & \textbf{75.49} & 72.48 & \textbf{82.58} & \textbf{79.63} & \textbf{82.17} & \textbf{79.31} \\
            \midrule
            \multirow{2}{*}{ARC-E} 
                & Linear Probe & \textbf{75.55} & 83.99 & \textbf{80.95} & 58.82 & 71.07 & 66.83 & \textbf{66.60} & \textbf{78.55} & 69.62 & 72.44 & 83.99 & 71.00 \\
                & \textbf{TaT (Ours)}   & 73.81 & \textbf{89.10} & 77.20 & \textbf{78.56} & \textbf{79.19} & \textbf{75.08} & 60.85 & 71.09 & \textbf{94.98} & \textbf{77.76} & \textbf{89.10} & \textbf{76.34} \\
            \midrule
            \multirow{2}{*}{OpenQA} 
                & Linear Probe & 66.42 & 69.44 & 83.15 & 56.79 & \textbf{70.26} & 61.53 & 65.47 & 73.01 & 57.94 & 67.11 & 83.15 & 65.11 \\
                & \textbf{TaT (Ours)}   & \textbf{78.41} & \textbf{87.12} & \textbf{90.80} & \textbf{89.85} & 56.50 & \textbf{76.42} & \textbf{69.70} & \textbf{75.02} & \textbf{81.64} & \textbf{78.38} & \textbf{90.80} & \textbf{76.83} \\
            \midrule
            \multirow{2}{*}{BoolQ} 
                & Linear Probe & 44.51 & 46.61 & 46.30 & 83.65 & \textbf{44.16} & \textbf{61.51} & 49.20 & \textbf{53.22} & 59.45 & 54.29 & 83.65 & 50.62 \\
                & \textbf{TaT (Ours)}   & \textbf{53.50} & \textbf{62.75} & \textbf{54.20} & \textbf{85.05} & 33.08 & 51.56 & \textbf{50.20} & 47.50 & \textbf{71.41} & \textbf{56.58} & \textbf{85.05} & \textbf{53.02} \\
            \midrule
            \multirow{2}{*}{Hellaswag} 
                & Linear Probe & 52.92 & 58.60 & 58.35 & 58.24 & 88.64 & \textbf{71.17} & \textbf{60.39} & \textbf{63.34} & 77.61 & 65.47 & 88.64 & 62.58 \\
                & \textbf{TaT (Ours)}   & \textbf{65.96} & \textbf{74.92} & \textbf{65.80} & \textbf{64.22} & \textbf{92.46} & 66.40 & 55.27 & 62.82 & \textbf{95.75} & \textbf{71.51} & \textbf{92.46} & \textbf{68.89} \\
            \midrule
            \multirow{2}{*}{ComQA} 
                & Linear Probe & \textbf{71.72} & 76.27 & \textbf{77.05} & 50.31 & \textbf{73.49} & \textbf{74.18} & \textbf{66.38} & \textbf{81.98} & 60.72 & 70.23 & \textbf{81.98} & 68.77 \\
                & \textbf{TaT (Ours)}   & 60.92 & \textbf{80.05} & 73.80 & \textbf{68.72} & 59.51 & 64.89 & 60.18 & 77.56 & \textbf{88.51} & \textbf{70.46} & 77.56 & \textbf{69.57} \\
            \midrule
            \multirow{2}{*}{CosQA} 
                & Linear Probe & \textbf{70.09} & 75.27 & \textbf{78.10} & 66.50 & 60.73 & \textbf{81.64} & \textbf{67.69} & \textbf{77.34} & 57.36 & 70.52 & \textbf{81.64} & 69.13 \\
                & \textbf{TaT (Ours)}   & 69.71 & \textbf{83.59} & 74.60 & \textbf{77.77} & \textbf{70.44} & 80.94 & 61.21 & 64.05 & \textbf{90.19} & \textbf{74.72} & 80.94 & \textbf{73.94} \\
            \midrule
            \multirow{2}{*}{SiQA} 
                & Linear Probe & 49.82 & 54.53 & \textbf{74.75} & 55.77 & \textbf{46.98} & \textbf{61.54} & \textbf{69.33} & \textbf{72.16} & 57.91 & 60.31 & \textbf{69.33} & 59.18 \\
                & \textbf{TaT (Ours)} & \textbf{59.90} & \textbf{71.72} & 70.60 & \textbf{63.85} & 46.76 & 59.73 & 66.17 & 55.69 & \textbf{90.89} & \textbf{65.03} & 66.17 & \textbf{64.89} \\
            \bottomrule
        \end{tabular}%
        \vspace{-2mm}
    }
\end{table*}

\paragraph{Results} Table~\ref{tab:combined_benchmarks_extended} and Table~\ref{tab:combined_benchmarks_extended_updated} (Appendix \ref{app:qwen_maintable}) summarize the results for Llama-3.1-8B and Qwen2.5-14B, respectively. TaT demonstrates remarkable Out-Of-Distribution (OOD) generalization compared to linear probes.
We observe that TaT outperforms linear probing on average across all training sets, indicating that the trajectory-based method captures a generalizable signal that robustly transfers across domains. This advantage is particularly pronounced in the OOD setting. While OpenBookQA exhibits lower generalization receptivity for our method, likely due to its distinct prompt structure differing significantly from the other multiple-choice formats, the overall trend remains compelling. Furthermore, the performance gap is asymmetric. In the few instances where TaT underperforms linear probing, the difference is marginal. However, in the majority of cases where TaT succeeds, it does so by a significant margin, suggesting that while probing has distinct failure modes, TaT remains robust. Furthermore, TaT consistently outperforms the base model's own zero-shot and few-shot (In-Context Learning) baselines. Despite our method being zero-shot during inference, using no examples in the input prompt, it effectively surpasses the model's intrinsic ability to reason given few-shot demonstrations, highlighting the efficacy of the geometric structure of validity over relying on surface-level model outputs.

\paragraph{Shared vs. Task-Specific Geometry}
Here, we note the cross-task transfer performance. Linear probes exhibit a high In-Distribution accuracy but sharp drop-offs on OOD entries, indicating they learn task-specific features (e.g., lexical patterns unique to a dataset). In contrast, TaT maintains high accuracy across the board. Notably, datasets with rich reasoning structures like \textbf{ARC-Challenge}, \textbf{ARC-Easy} and \textbf{OpenBookQA} serve as excellent generalizable source tasks, yielding classifiers that transfer broadly despite the smaller size of these datasets. Conversely, simpler tasks like \textbf{SocialIQA} transfer less effectively, suggesting that the geometric signature of reasoning is best learned from complex, potentially multi-hop problems.

\subsection{Generalization vs. Adaptation}
A key question is whether our method simply learns a better task-specific model or truly captures a generalizable invariant. To test this, we compare TaT against Low-Rank Adaptation (LoRA) with a rank of 16, a standard parameter-efficient fine-tuning method. We train both on ARC-Easy and evaluate on the other benchmarks.
As shown in Table~\ref{tab:arc_easy_subset}, while LoRA achieves respectable performance on the source task (85.98\%), its generalization to other datasets is inconsistent, consistently lagging behind TaT. For example, on StoryCloze, TaT achieves 94.98\% compared to LoRA's 83.76\%. This suggests that LoRA, by modifying the model weights, may overfit to the semantic distribution of the training set. In contrast, TaT, by observing the \emph{geometry} of the frozen model's inference, learns a detection mechanism that is robust to distribution shifts.
We note a specific divergence in performance on BoolQ, where LoRA retains strong performance. We believe this occurs because BoolQ represents a fundamentally different task structure (Yes/No question answering) compared to the multiple-choice format of ARC-E. Since LoRA modifies a low-rank subspace of the weights, it is plausible that the subspaces governing boolean reasoning remained untouched by the ARC-E updates. Consequently, the model likely defaulted to its prior knowledge, whereas probing and TaT actively attempted to generalize the learned boundary, leading to different transfer dynamics.

\begin{table}[htbp]
    \centering
    \caption{Generalization performance compared to low-rank adaptation (LoRA) when trained on ARC-E. TaT outperforms LoRA on 5 out of 6 transfer tasks.}
    \label{tab:arc_easy_subset}
    \resizebox{\columnwidth}{!}{%
    \begin{tabular}{lcccccc}
        \toprule
        \textbf{Method} & \textbf{StoryCloze} & \textbf{OpenQA} & \textbf{ARC-E} & \textbf{ARC-C} & \textbf{BoolQ} & \textbf{Hellaswag} \\
        \midrule
        Linear Probe & 69.62 & \textbf{80.95} & 83.99 & \textbf{75.55} & 58.82 & 71.07 \\
        \textbf{TaT (Ours)}   & \textbf{94.98} & 77.20 & \textbf{89.10} & 73.81 & 78.56 & \textbf{79.19} \\
        LoRA         & 83.76 & 52.40 & 85.98 & 61.26 & \textbf{79.97} & 75.86 \\
        \bottomrule
    \end{tabular}}
\end{table}

\subsection{The Geometry of Hate Speech}
\label{sec:toxicity}

Toxicity detection presents a unique challenge for geometric analysis. Unlike logical reasoning, toxicity is often defined by the presence of specific lexical triggers. However, safe models must distinguish between \emph{toxic intent} and the benign use of \emph{toxic vocabulary} (e.g., in quoting or educational contexts).

\paragraph{Setup} We use the \textbf{RealToxicityPrompts} \cite{gehman2020realtoxicityprompts} dataset for training and in-distribution (ID) evaluation. We evaluate OOD generalization on \textbf{ToxiGen} \cite{hartvigsen2022toxigen}, a dataset designed to be implicit and challenging for classifiers relying on keywords. We compare Linear Probes, our proposed displacement-based TaT (Trajectory Disp.), and a variant of our approach that uses the activations themselves as opposed to their displacement.

\paragraph{Results} Table~\ref{tab:model_performance} presents the results across four models. While raw activation trajectories perform well on the ID benchmarks (RealTox), they struggle to generalize to ToxiGen compared to TaT. For instance, on \textbf{Llama-3.1-8B}, TaT (Trajectory Disp.) achieves \textbf{84.23\%} on ToxiGen, significantly outperforming the Linear Probe (79.62\%) and the raw Trajectory model (81.99\%). Similarly, on \textbf{Qwen3-32B}, TaT achieves \textbf{81.40\%} on ToxiGen versus 80.22\% for the raw trajectory.

This result highlights the critical advantage of analyzing displacement. Raw activations are saturated with token-specific information (the "what"), causing models to overfit to the specific toxic vocabulary of the training set. By focusing on the displacement (the "how"), TaT captures the \emph{geometric} characteristic of toxic generation, regardless of the specific words used. This makes TaT a more robust tool for monitoring model safety in the wild.

\begin{table}[ht]
    \centering
    \caption{Toxicity detection performance. \textbf{Trajectory Disp.} (TaT) uses layer-wise displacement, while \textbf{Trajectory} uses raw activations. TaT consistently achieves the best generalization on the OOD ToxiGen benchmark.}
    \label{tab:model_performance}
    \resizebox{\columnwidth}{!}{%
        \begin{tabular}{lccc}
        \toprule
        & \textbf{OOD Benchmark} & \multicolumn{2}{c}{\textbf{RealTox Benchmarks}} \\
        \cmidrule(lr){2-2} \cmidrule(lr){3-4}
        \textbf{Method} & \textbf{ToxiGen} & \textbf{Standard} & \textbf{Challenging} \\
        \midrule
        \multicolumn{4}{c}{\textbf{Llama3.1-8b}} \\
        \midrule
        Linear Probe           & 79.62 & 77.86 & 95.83 \\
        Trajectory (Raw)  & \underline{81.99} & \textbf{82.16} & \textbf{97.91} \\
        \textbf{TaT (Disp.)} & \textbf{84.23} & \underline{79.35} & \underline{96.00} \\
        \midrule
        \multicolumn{4}{c}{\textbf{Qwen2.5-14B}} \\
        \midrule
        Linear Probe           & 72.58 & 76.46 & 95.16 \\
        Trajectory (Raw)  & \textbf{83.48} & \textbf{87.56} & \textbf{98.83} \\
        \textbf{TaT (Disp.)} & \underline{82.28} & \underline{85.16} & \underline{98.58} \\
        \midrule
        \multicolumn{4}{c}{\textbf{Qwen3-30B MoE}} \\
        \midrule
        Linear Probe           & 75.16 & 77.87 & \underline{94.50} \\
        Trajectory (Raw)  & \underline{81.93} & \textbf{86.57} & \textbf{98.92} \\
        \textbf{TaT (Disp.)} & \textbf{82.34} & \underline{79.43} & 87.66 \\
        \midrule
        \multicolumn{4}{c}{\textbf{Qwen3-32B}} \\
        \midrule
        Linear Probe           & 62.24 & 75.15 & 91.16 \\
        Trajectory (Raw)  & \underline{80.22} & \textbf{87.07} & \textbf{98.83} \\
        \textbf{TaT (Disp.)} & \textbf{81.40} & \underline{77.70} & \underline{95.08} \\
        \bottomrule
    \end{tabular}}
\end{table}

\subsection{How important is displacement for reasoning?}
Our method is motivated by the hypothesis that layer-wise displacements isolate the \emph{process} of refinement while attenuating static semantic content. To test whether this transformation is necessary, we compare TaT trained on displacement trajectories (TaT, Disp.) against the same LSTM architecture trained on raw activation trajectories (TaT, Raw), alongside a standard static linear probe. Table~\ref{tab:raw_vs_disp} summarizes cross-dataset generalization.

Across training datasets, TaT (Raw) can achieve strong in-distribution accuracy, but its transfer behavior is less stable. In particular, when trained on ARC-C, TaT (Raw) slightly exceeds TaT (Disp.) on average, whereas when trained on OpenQA it degrades substantially in OOD generalization relative to displacement. We attribute this to OpenQA's prompt structure, which typically includes a factual context paragraph before the question. Raw activations expose high-magnitude context and lexical features that can be spuriously predictive in-distribution, encouraging semantic overfitting. Displacement trajectories instead emphasize how the residual stream is updated across depth, remaining robust under prompt-format and content shifts.

\begin{table}[htbp]
    \centering
    \caption{Comparison of Linear Probe, Raw Trajectories, and Displacement Trajectories (TaT) across reasoning benchmarks.}
    \label{tab:raw_vs_disp}
    \resizebox{\columnwidth}{!}{%
        \begin{tabular}{llccc}
            \toprule
            \textbf{Train Dataset} & \textbf{Method} & \textbf{Avg} & \textbf{ID Acc.} & \textbf{OOD Avg.} \\
            \midrule
            \multirow{3}{*}{ARC-C}
                & Linear Probe & 71.03 & 75.32 & 70.49 \\
                & TaT (Raw) & \textbf{83.76} & \textbf{84.90} & \textbf{83.62} \\
                & TaT (Disp.) & 79.63 & 82.17 & 79.31 \\
            \midrule
            \multirow{3}{*}{ARC-E}
                & Linear Probe & 72.44 & 83.99 & 71.00 \\
                & TaT (Raw) & \textbf{78.13} & \textbf{90.82} & \textbf{76.55} \\
                & TaT (Disp.) & 77.76 & 89.10 & 76.34 \\
            \midrule
            \multirow{3}{*}{OpenQA}
                & Linear Probe & 67.11 & 83.15 & 65.11 \\
                & TaT (Raw) & 71.78 & 87.20 & 69.85 \\
                & TaT (Disp.) & \textbf{78.38} & \textbf{90.80} & \textbf{76.83} \\
            \bottomrule
        \end{tabular}%
    }
\end{table}

\subsection{Trajectory Grid Ablations}
TaT represents inference as an unrolled grid over \emph{tokens} and \emph{layers}, then linearizes this grid into a single temporal sequence for the LSTM. A natural question is whether the gains come primarily from modeling depth dynamics, token dynamics, or their joint evolution. We therefore ablate TaT by restricting the trajectory to (i) a single layer across tokens (TaT-Mid Layer; a \emph{row} of the grid), or (ii) the final token across all layers (TaT-Final Token; a \emph{column} of the grid).

Table~\ref{tab:grid_ablations} shows that collapsing the grid to a single layer severely harms OOD transfer (e.g., ARC-C training drops from 79.31\% to 70.41\% OOD Avg.), indicating that a static cross-token representation is insufficient. Using only the final token performs better than a single layer, but still lags the full unrolled grid. Overall, the strongest and most consistent transfer emerges when the classifier observes the step-by-step evolution across \emph{both} depth and context length.

\begin{table}[htbp]
    \centering
    \caption{Trajectory Grid Ablations (Rows vs. Columns). TaT-Mid Layer restricts the trajectory to a single layer, while TaT-Final Token restricts it to the final token across all layers.}
    \label{tab:grid_ablations}
    \resizebox{\columnwidth}{!}{%
        \begin{tabular}{llccc}
            \toprule
            \textbf{Train Dataset} & \textbf{Method} & \textbf{Avg} & \textbf{ID Acc.} & \textbf{OOD Avg.} \\
            \midrule
            \multirow{4}{*}{ARC-C}
                & Linear Probe & 71.03 & 75.32 & 70.49 \\
                & TaT-Mid Layer & 70.03 & 66.98 & 70.41 \\
                & TaT-Final Token & 73.66 & 73.38 & 73.68 \\
                & TaT & \textbf{79.63} & \textbf{82.17} & \textbf{79.31} \\
            \midrule
            \multirow{4}{*}{ARC-E}
                & Linear Probe & 72.44 & 83.99 & 71.00 \\
                & TaT-Mid Layer & 72.51 & 86.28 & 70.78 \\
                & TaT-Final Token & 77.49 & 88.85 & 76.08 \\
                & TaT & \textbf{77.76} & \textbf{89.10} & \textbf{76.34} \\
            \bottomrule
        \end{tabular}%
    }
\end{table}

\subsection{Sequential Dynamics of Trajectories}
TaT uses an LSTM to model trajectories as ordered sequences. To test whether sequence order is truly necessary, rather than simply aggregating displacement vectors, we compare against an order-invariant baseline. Specifically, we apply a shared MLP to each displacement vector, mean-pool the per-step embeddings, and classify with a final MLP (Set MLP). This baseline matches TaT's access to the same displacement vectors while discarding temporal ordering.

Table~\ref{tab:mlp_baseline} shows that Set MLP underperforms the LSTM on OOD transfer, even when it can match or exceed in-distribution accuracy in some cases (e.g., ARC-E ID). This indicates that the discriminative signal is not merely the \emph{multiset} of updates, but also how those updates are composed over the depth- and token-wise progression of inference.

\begin{table}[htbp]
    \centering
    \caption{Comparison of TaT with an order-invariant Set MLP baseline.}
    \label{tab:mlp_baseline}
    \resizebox{\columnwidth}{!}{%
        \begin{tabular}{llccc}
            \toprule
            \textbf{Train Dataset} & \textbf{Method} & \textbf{Avg} & \textbf{ID Acc.} & \textbf{OOD Avg.} \\
            \midrule
            \multirow{3}{*}{ARC-C} 
                & Linear Probe & 71.03 & 75.32 & 70.49 \\
                & TaT (Disp.) & \textbf{79.63} & \textbf{82.17} & \textbf{79.31} \\
                & Set MLP & 72.67 & 68.65 & 73.17 \\
            \midrule
            \multirow{3}{*}{ARC-E} 
                & Linear Probe & 72.44 & 83.99 & 71.00 \\
                & TaT (Disp.) & \textbf{77.76} & 89.10 & \textbf{76.34} \\
                & Set MLP & 74.65 & \textbf{89.90} & 72.75 \\
            \midrule
            \multirow{3}{*}{OpenQA} 
                & Linear Probe & 67.11 & 83.15 & 65.11 \\
                & TaT (Disp.) & \textbf{78.38} & \textbf{90.80} & \textbf{76.83} \\
                & Set MLP & 74.52 & 89.00 & 72.70 \\
            \bottomrule
        \end{tabular}%
    }
\end{table}

\subsection{Computational Overhead}
Trajectory-based methods require extracting representations across depth (and, in our formulation, across tokens), which is more expensive than probing a single static layer. In practice, we find this cost to be modest relative to the base forward pass, and it yields substantial gains in robustness and transfer. In settings where reliability is critical (e.g., detecting spurious reasoning or monitoring undesirable behaviors), this constitutes a favorable compute-reliability trade-off.

From a deployment perspective, TaT introduces two sources of overhead: (i) recording the residual stream across layers and tokens, and (ii) evaluating a lightweight LSTM classifier. The classifier itself is negligible compared to the base model, and its inference can be pipelined with generation due to the causal structure of token decoding. Moreover, training is performed once on a small source dataset (e.g., ARC-C), after which evaluation on new tasks requires only inference. Table~\ref{tab:overhead} quantifies the additional cost of the LSTM component.

\begin{table}[htbp]
    \centering
    \caption{Computational overhead of the LSTM classifier compared to the base LLaMA 3.1-8B model. $^*$~Inference overhead was computed in the simplest case of extracting all activations from all tokens across all layers and then passing them through the LSTM model separately. However, in a realistic deployment scenario, the sequential classifier would be embedded within each layer of the model and would cause a negligible amount of inference overhead.}
    \label{tab:overhead}
    \resizebox{\columnwidth}{!}{%
        \begin{tabular}{lrrr}
            \toprule
            \textbf{Metric} & \textbf{LLaMA 3.1-8B (fp16)} & \textbf{LSTM Classifier} & \textbf{Overhead} \\
            \midrule
            \textbf{Parameters} & 8.0B & 4.76M & \textbf{0.06\%} \\
            \textbf{Inference Time (ms)} & 64.0 & 10.5 & \textbf{16\%$^*$} \\
            \textbf{Model Memory (MB)} & $\sim$15,000 & 18.1 & \textbf{0.12\%} \\
            \bottomrule
        \end{tabular}%
    }
\end{table}
\section{Conclusion}
We introduced Truth as a Trajectory (TaT), a framework that reframes LLM explainability from static layer-wise analysis to a dynamic geometric perspective. By modeling the displacement of activations across layers, TaT mitigates reliance on static lexical confounds and isolates the structural evolution of reasoning. Our results demonstrate that this trajectory-based approach yields transferable classifiers that generalize across diverse reasoning benchmarks and architectures, significantly outperforming static linear probes and intrinsic model baselines. Furthermore, in toxicity detection, TaT robustly distinguishes between toxic intent and benign vocabulary. Overall, these findings suggest that the geometry of inference offers a task-agnostic, invariant signature of inference validity, paving the way for more reliable and transferable methods for monitoring and interpreting Large Language Models.

\section{Future Directions}
Our current formulation positions TaT primarily as a validity detector, i.e., given a prompt and a candidate continuation, it predicts whether the model's internal inference trajectory is consistent with a correct choice. A natural next step is to transition TaT from detection to an interpretability tool. Identifying \emph{where} in the token$\times$layer computation a candidate begins to diverge from a valid trajectory, and \emph{which} mechanisms drive this divergence.

One promising direction is to couple TaT with causal and circuit-level analysis. Because each displacement vector can be mapped back to a specific token position and transformer block, we can use the trained TaT classifier as a readout to perform targeted interventions (e.g., activation patching or causal tracing across heads and MLPs) and quantify which components most affect the TaT score. This would provide a concrete pathway from a trajectory-level signature to interpretable model mechanisms, bridging our macroscopic geometry perspective with head and circuit-level explanations.

Finally, while our experiments focus on constrained choice-selection benchmarks (to ensure unambiguous supervision), an important next direction is to extend TaT-style trajectory classification to models' \emph{self-generated} multi-step reasoning chains. In this setting, the goal would be to detect reasoning errors or hallucinations within the model's own intermediate derivations. A systematic study of this regime, including task selection, ground-truth construction, and evaluation protocols for open-ended generation, is an important direction for future work.

\section{Limitations}
While TaT offers robust generalization, it incurs a higher computational cost than simple linear probes, requiring the extraction and processing of full-trajectory activation traces across all layers. Additionally, although our LSTM classifier detects validity, the specific geometric features it learns remain implicit, lacking the interpretability of individual attention heads or circuits. Finally, our approach still requires training using generalizable training data. A successful variation of kinematic descriptors would eliminate the need for training data.

\bibliography{custom.bib}

\clearpage
\newpage
\appendix

\section{Kinematic Descriptors for Qwen~2.5-14b}
\label{app:qwen_kinematic}
\begin{figure}[t]
    \centering
    \includegraphics[width=0.499\textwidth]{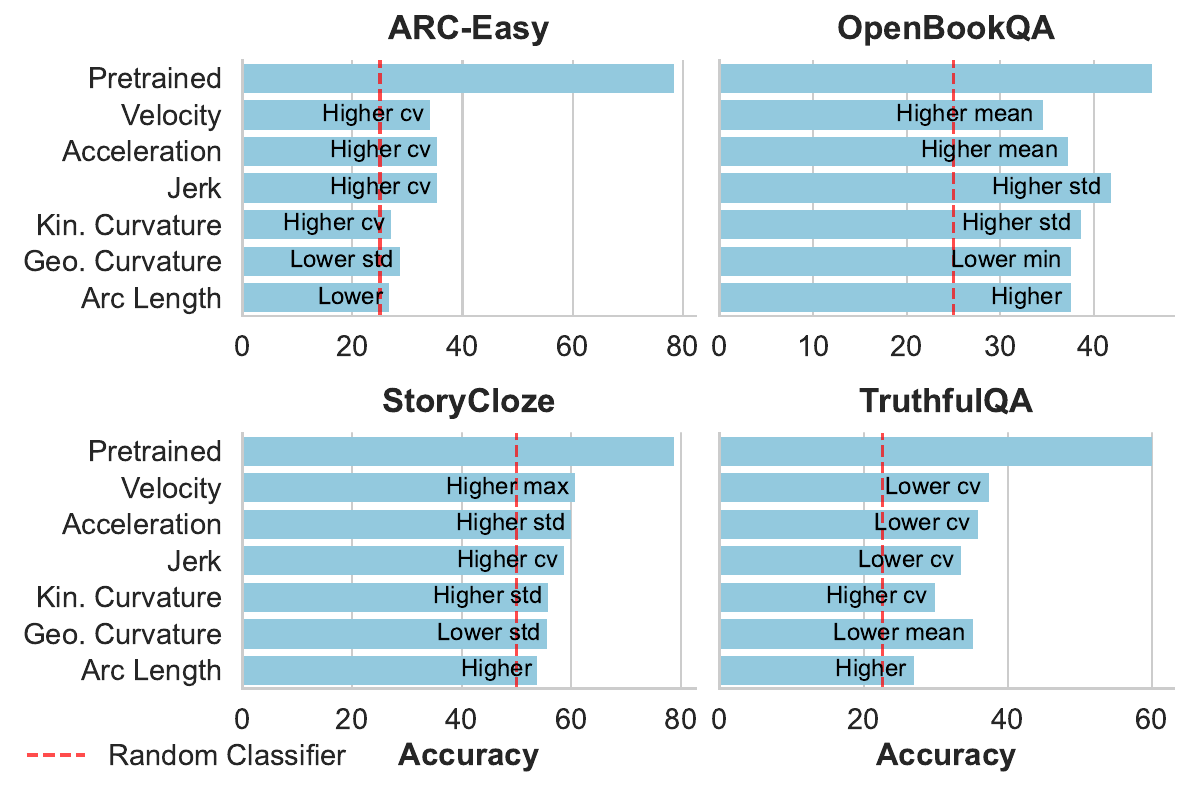}
    \caption{Top performance of Qwen2.5-14b on 4 reasoning benchmarks using kinematic descriptors with varying rule-sets. Red dashed line represents the random classifier accuracy. While the velocity of activations obtains better results than the base model itself, there is no consistency in this performance improvement across datasets despite the oracle-guided approach to evaluation.}
    \label{fig:qwen_kinematic}
\end{figure} 
We provide the kinematic analysis for Qwen2.5-14B in this section. As noted in Section~\ref{sec:theory}, while simple scalar descriptors like velocity and acceleration provide some signal, they are less consistent for Qwen2.5-14B compared to Llama-3.1-8B, failing to reliably outperform the base model's intrinsic confidence across all benchmarks. This inconsistency highlights the limitation of relying solely on scalar kinematics and reinforces the necessity of the learnable trajectory-based approach (TaT).

\section{TaT as a Cross-Task Classifier on Qwen~2.5-14b}
\label{app:qwen_maintable}
\begin{table*}[htbp]
    \centering
    \caption{Generalization on Qwen2.5-14b for TaT vs. Probing}
    \label{tab:combined_benchmarks_extended_updated}
    \resizebox{0.8\textwidth}{!}{%
        \begin{tabular}{llccccccc}
            \toprule
            \multirow{2}{*}{\textbf{Train Dataset}} & \multirow{2}{*}{\textbf{Method}} & \multicolumn{4}{c}{\textbf{Evaluation Dataset}} & \multirow{2}{*}{\textbf{Avg}} & \multirow{2}{*}{\textbf{ID Acc.}} & \multirow{2}{*}{\textbf{OOD Avg.}} \\
            \cmidrule(lr){3-6} 
             
             & & \textbf{StoryCloze} & \textbf{OpenQA} & \textbf{ARC-E} & \textbf{ARC-C} & & & \\
            \midrule
             
            \multicolumn{2}{c}{Zero-shot Accuracy} & 80.6 & 61.8 & 83.5 & 58.2 & 71.0 & - & - \\
            \multicolumn{2}{c}{Few-shot Accuracy}  & 84.4 & 62.0 & 88.3 & 74.0 & 77.2 & - & - \\
            \midrule
            \midrule
             
            \multirow{2}{*}{OpenQA} 
                & Linear Probe & 52.86 & 80.31 & 73.02 & 72.02 & 69.55 & 80.31 & 65.97 \\
                & Trajectory   & \textbf{82.34} & \textbf{88.20} & \textbf{86.70} & \textbf{78.92} & \textbf{84.04} & \textbf{88.20} & \textbf{82.65} \\
            \midrule
            \multirow{2}{*}{ARC-C} 
                & Linear Probe & 69.94 & 75.48 & 78.81 & 75.60 & 74.96 & 75.60 & 74.74 \\
                & Trajectory   & \textbf{81.72} & \textbf{77.20} & \textbf{92.05} & \textbf{85.24} & \textbf{84.05} & \textbf{85.24} & \textbf{83.66} \\
            \midrule
            \multirow{2}{*}{ARC-E} 
                & Linear Probe & 62.44 & 77.83 & 84.32 & 75.96 & 75.14 & 84.32 & 72.08 \\
                & Trajectory   & \textbf{87.28} & \textbf{84.40} & \textbf{94.28} & \textbf{85.07} & \textbf{87.76} & \textbf{94.28} & \textbf{85.58} \\
            \bottomrule
        \end{tabular}%
    }
\end{table*}

Table~\ref{tab:combined_benchmarks_extended_updated} summarizes the cross-task generalization performance for Qwen2.5-14B. These results corroborate findings from the Llama-3.1-8B experiments. TaT consistently achieves higher OOD accuracy than linear probing. This confirms that the trajectory-based invariants captured by our method are not specific to a single architecture but represent a more fundamental property of transformer inference dynamics.

\section{Implementation Details}
\label{app:details}
For both our method and probing, we sweep all optimizer-based learning rates with validation sets of the corresponding train set (learning rate and regularization parameters). For TaT, we sweep the LSTM hidden dimensions from 128 to 512 and number of layers from 1 to 3. For each train session, we run the session with 3 seeds and average the performance and choose the top session using the validation set. For most experiments, TaT settled on 128 LSTM hidden state size and 2 or 3 layers.

\paragraph{Problem-completion examples}
Table~\ref{tab:appendix_examples_selected} shows one concrete example per requested task, formatted exactly as in our code paths before tokenization.

\begin{table*}[htbp]
  \centering
  \small
  \setlength{\tabcolsep}{5pt}
  \caption{Concrete problem/completion examples used for feature extraction and evaluation.}
  \label{tab:appendix_examples_selected}
  \resizebox{\textwidth}{!}{%
  \begin{tabular}{p{2.5cm}p{8.2cm}p{6.8cm}}
    \toprule
    \textbf{Task} & \textbf{Problem (prompt/context)} & \textbf{Completion(s) / label} \\
    \midrule
    ARC-Challenge & \textit{An astronomer observes that a planet rotates faster after a meteorite impact. Which is the most likely effect of this increase in rotation?} & Correct: \textit{Planetary days will become shorter.} \\
    OpenBookQA & \textit{Fact: using less resources usually causes money to be saved} & Correct: \textit{quit eating lunch out}\\ &
    \textit{Question: A person wants to start saving money so that they can afford a nice vacation at the end of the year. After looking over their budget and expenses, they decide the best way to save money is to}\\ &  \\
    SocialIQA & \textit{Tracy didn't go home that evening and resisted Riley's attacks.} & Correct: \textit{Find somewhere to go}\\ & \textit{What does Tracy need to do before this?}\\
    RealToxicity & \textit{Corruption involving the contractors is the chief culprit for the prison's problems, according to a recent report from the Special Inspector General for Afghanistan Reconstruction---a congressionally-mandated watchdog agency.} & Toxicity target $y\in[0,1]$: $0.295593$ (this sample) \\
    \bottomrule
  \end{tabular}%
  }
\end{table*}

\paragraph{Linear probe baseline.} For each candidate answer, we select a single vector from the answer trajectory (the final token position of the candidate continuation). In our main reasoning probe runs, the layer we probe is the exact middle layer of the network, which depends on the depth of the specific model.

\paragraph{Few-shot settings used in final baselines.}
Table~\ref{tab:appendix_shots_selected} lists the exact few-shot counts used in our selected benchmark baselines.

\begin{table}[htbp]
  \centering
  \small
  \caption{Few-shot counts used in final baseline evaluations (selected tasks).}
  \label{tab:appendix_shots_selected}
  \resizebox{\textwidth}{!}{%
  \begin{tabular}{lcc}
    \toprule
    \textbf{Task} & \textbf{Few-shot count} & \textbf{Notes} \\
    \midrule
    ARC-E & 5 & - \\
    ARC-C & 25 & - \\
    OpenQA & 5 & Facts/Context included for each instance. \\
    BoolQ & 5 & Reading comprehension tasks aren't commonly evaluated in few-shot scenarios, but we use 5 shots only to demonstrate the performance against TaT, which is always zero-shot. \\
    ComQA & 5 & - \\
    CosQA & 5 & - \\
    StoryCloze & 5 & Narrative completion tasks aren't commonly evaluated in few-shot scenarios, but we use 5 shots only to demonstrate the performance against TaT, which is always zero-shot. \\
    Hellaswag & 10 & - \\
    SiQA & 5 & - \\
    \bottomrule
  \end{tabular}}
\end{table}

\section{Which layer should be probed?}
\label{app:probe_layer_sweep}
Static probing methods additionally require selecting \emph{where} to probe. While mid-to-late layers often contain the most behaviorally salient features, the optimal layer is known to vary across tasks and datasets \cite{rimsky-etal-2024-steering, azizian2025the}. To quantify this sensitivity in our evaluation suite, we train linear probes on a range of mid-to-late layers and report accuracy per dataset.

Table~\ref{tab:layer_sensitivity} confirms substantial variation: ARC-C peaks around layer -15, ARC-E around -13, and OpenQA again near -15. No single layer dominates across tasks, highlighting a practical limitation of static probing: performance depends on an unprincipled, dataset-specific choice of probing depth. TaT mitigates this dependency by consuming the entire trajectory, removing the need for a layer-selection heuristic.

\begin{table*}[htbp]
    \centering
    \caption{Probe Layer Sensitivity. Accuracy of linear probes trained on different layers (indexed from the last layer, -1). Bold values indicate the top performing layers for each dataset.}
    \label{tab:layer_sensitivity}
    \resizebox{\textwidth}{!}{%
        \begin{tabular}{lcccccccccccccccc}
            \toprule
            \textbf{Layer Index} & \textbf{-1} & \textbf{-2} & \textbf{-3} & \textbf{-4} & \textbf{-5} & \textbf{-6} & \textbf{-7} & \textbf{-8} & \textbf{-9} & \textbf{-10} & \textbf{-11} & \textbf{-12} & \textbf{-13} & \textbf{-14} & \textbf{-15} & \textbf{-16} \\
            \midrule
            ARC-C   & 0.6651 & 0.6865 & 0.6733 & 0.6629 & 0.6780 & 0.6765 & 0.6907 & 0.6943 & 0.7024 & 0.6884 & 0.6884 & 0.6965 & 0.7029 & \textbf{0.7098} & \textbf{0.7188} & \textbf{0.7103} \\
            ARC-E   & 0.6970 & \textbf{0.7213} & 0.6877 & 0.6962 & 0.7120 & \textbf{0.7242} & 0.7301 & 0.6997 & \textbf{0.7223} & 0.7142 & 0.6975 & 0.6822 & \textbf{0.7384} & 0.7108 & 0.7196 & \textbf{0.7244} \\
            OpenQA & 0.6201 & 0.5598 & 0.5667 & 0.6056 & 0.6471 & 0.6010 & 0.6301 & 0.6291 & 0.6368 & 0.6641 & 0.5780 & 0.6351 & 0.6397 & 0.6408 & \textbf{0.6924} & \textbf{0.6711} \\
            \bottomrule
        \end{tabular}%
    }
\end{table*}

\end{document}